\documentclass[11pt]{article}
\usepackage[margin=1in]{geometry}
\usepackage{amsmath,amssymb}
\usepackage{booktabs}
\usepackage{graphicx}
\usepackage[square,authoryear]{natbib}
\usepackage[activate=false]{microtype}
\usepackage{listings}
\usepackage{xcolor}
\usepackage[colorlinks=true,linkcolor=blue!60!black,citecolor=blue!60!black,urlcolor=blue!60!black]{hyperref}
\usepackage{xurl}

\providecommand{\tightlist}{\setlength{\itemsep}{2pt}\setlength{\parskip}{0pt}}
\title{Retrofitting Recurrent Depth into a Pretrained Language Model:\\Installation, Extrapolation, Transfer, and Retention at Two Parameter Budgets}
\author{Mark Shapiro\\\normalsize ORCID: \href{https://orcid.org/0000-0002-7611-5432}{0000-0002-7611-5432}}
\date{}
\begin{document}
\maketitle
\begin{abstract}
A dense, pretrained language model can be retrofitted with recurrent depth and learn an iterative latent transition that persists after outcome-only annealing. Qwen2.5-0.5B-Instruct is split into a Prelude, a weight-tied Recurrent Block, and a Coda, with an identity-preserving one-loop path and a trainable bridge that re-injects the Prelude representation on later loops. At loop 1 the retrofit remains non-inferior to its base on a preregistered ARC battery.

Three findings. First, the mechanism is a reusable procedure rather than terminal-answer lookup, and installs at two budgets: 6M trained parameters over frozen base weights and 180M full-block. With intermediate-step supervision, the model computes one task step per loop, persists when only final answers are graded, and keeps stepping past a problem\textquotesingle s target depth. The adapter matched the full block overall (83.8\% versus 84.0\%), led through depth 11, and trailed beyond. With verbal fine-tuning, the full-block variant reached 79-86\% on controlled verbal renderings of the task, while zero-shot transfer was minimal at both budgets, and adapter verbal training begun from the installed mechanism outpaced matched fresh training by 18.6 points, including on a held-out test set. Second, the operation extrapolates to roughly 1.5 times its supervised depth, holding 70\% accuracy through depth 18. Third, a same-size model fine-tuned to write its reasoning out in tokens matched the recurrent model within its learned horizon but collapsed beyond it. The recurrent model won overall, 84\% versus 72\%, retained 53\% versus 2.5\% beyond depth 10, and answered 7.6 times faster in fewer output tokens. This system-level comparison demonstrates that an iterative transformer can perform deeper reasoning in latent space faster than comparable or larger models fine-tuned on the same task.

A second training task, running the rule in reverse, exposed the limits: the inverse was learnable in isolation (63/64), but no continuation acquired it while preserving the installed mechanism and general capability, at either budget, a catastrophic-interference boundary. Learned depth selection remains an open problem for future investigation.
\end{abstract}

\hypertarget{introduction}{%
\section{Introduction}\label{introduction}}

Transformers normally obtain additional computation by adding layers during design or tokens during generation. Recurrent-depth models offer a third axis: repeatedly apply a shared transformation to hidden states. This can increase effective depth without increasing the number of distinct block parameters, and it creates a natural place for adaptive computation and latent iterative computation.

Part of this study\textquotesingle s motivation comes from small from-scratch recursive reasoners: namely the Hierarchical Reasoning Model (HRM) and the Tiny Recursive Model (TRM), which showed that iterated latent computation can enable networks with millions of parameters to perform deterministic reasoning tasks that defeat much larger models \citep{wang2025,jolicoeur2025}. Most recurrent-depth work, including that line, trains the recurrent architecture from scratch. The question here is whether the capability can be installed instead:

\begin{quote}\itshape
Can a pretrained, instruction-tuned dense language model be surgically converted into a recurrent-depth system, recover a stable iterative mechanism, and outperform registered dense recipes on tasks that reward depth, without damaging the base model\textquotesingle s general capability, and at what parameter cost?
\end{quote}

Because the repeated block receives states from a distribution different from the one it encountered during pretraining, looping alone tends to drive activations off manifold. The input context must therefore be re-injected on later iterations, and the pathway that combines it with the carried state must demonstrably receive training signal under the optimizer in use, a property that cannot be assumed. Furthermore, evaluation must be designed correctly to ensure that predictions are read at the correct point in the loop, on the surface that the model was trained to produce. The looping mechanism itself means that a correct state can be carried past its answer by further iteration. And finally, a consolidated mechanism can be erased by additional training.

Model conversion is therefore treated as a forensic engineering and measurement problem. The program uses exact identity checks, gradient-path audits with finite-difference cross-checks, forced-depth evaluations, frozen row manifests, same-reader scoring, paired statistical tests, checkpoint hashes, and guardrail hard stops. Negative results are retained and reported when they localize a boundary instead of being discarded as failed tuning runs.

The retrofit is studied at two training budgets over the identical surgery. Full-block adaptation of the looped region, 182,163,457 optimizer-marked and 180,556,929 forward-active parameters (Section 3.5), supplies the verbally trained transfer results and the benchmark preservation battery (Sections 10 and 7). The adapter budget, 6.01M forward-active parameters with the pretrained base weights untouched, supplies the installation floor, a matched depth profile, and its own persistence and retention results (Sections 6.2, 9.4, and 11.4). The two budgets frame the paper\textquotesingle s guiding comparison: whether the smaller intervention suffices for installing and operating the mechanism, and what the larger training budget buys. The answer is a measured crossover (Sections 6.3 and 9.4). The smaller budget suffices for installation and is slightly preferable through depth 11. The larger budget buys far-horizon extrapolation. The persistence signature and the retention boundary replicate at both budgets.

This paper is the deterministic study of a two-part program. A registered companion study addresses guided stochastic width on the substrate frozen here.

\subsection*{Contributions}

\begin{enumerate}
\def\labelenumi{\arabic{enumi}.}
\tightlist
\item
  \textbf{Two identity-preserving recurrent retrofits of Qwen2.5-0.5B over one surgery.} The model is split into Prelude, Recurrent Block, and Coda regions, and the one-loop route reproduces the base computation when recurrent additions are inactive. The identical surgery is trained at a full-block budget and at a frozen-base adapter budget (Section 6).
\item
  \textbf{Installation at two parameter budgets.} The mechanism installs under full-block adaptation, 180.6M forward-active, and under a rank-16 adapter configuration, 6.01M forward-active with base weights untouched.
\item
  \textbf{Evidence for persistent loop-indexed latent computation.} Exact intermediate-state supervision installs an iterative transition that survives outcome-only annealing and extends beyond its directly supported horizon, a constructive demonstration that a reusable procedure, and not only retrievable content, can be trained into a pretrained model.
\item
  \textbf{A support-scaling characterization.} The threshold-crossing validity frontier reaches 1.44-1.50 times the supervised support at training consolidation, at supports 4 through 12 across two alphabet sizes, with a measured tail ceiling rather than indefinite extrapolation.
\item
  \textbf{A registered dense-control comparison.} The recurrent system substantially exceeds direct and serialized-scratchpad dense recipes (the base model fine-tuned on the task directly, or fine-tuned to write out its logic in token space during the response) on identical frozen synthetic rows, especially beyond depth 10.
\item
  \textbf{A validation of preserved capability under loop installation, and the limits of that preservation.} Using a preregistered benchmark battery, the surgically modified recurrent Qwen is shown to be non-inferior on general capabilities at loop 1 across the training that installs the looping mechanism. The same battery detects real damage under other training regimes, defining where full-block continuation fails on this substrate, a boundary that replicated at the lower adapter budget.
\item
  \textbf{An optimizer-interaction finding.} Learning-rate control expressed as gradient scaling of one slice of a combined projection fails silently under both Adam-family and orthogonalizing optimizers, and the working solution, a separate tensor in a dedicated optimizer parameter group, is verified at the realized-update level (Section 5.1).
\end{enumerate}

\hypertarget{related-work}{%
\section{Related Work}\label{related-work}}

Related work falls into three lines: recurrent depth and adaptive computation, retrofits of pretrained transformers, and latent iterative computation in token and continuous space.

\hypertarget{recurrent-depth-and-adaptive-computation}{%
\subsection{Recurrent depth and adaptive computation}\label{recurrent-depth-and-adaptive-computation}}

Adaptive Computation Time lets recurrent networks learn how many internal steps to take before emitting an output \citep{graves2016}. Universal Transformers apply a self-attentive transition recurrently over depth and add dynamic per-position halting \citep{dehghani2018}. PonderNet formulates learned computation depth as a probabilistic halting process \citep{banino2021}. Deep Equilibrium Models solve directly for the fixed point of an effectively infinite weight-tied network \citep{bai2019}. Closer to the present regime, looped and recurrent-depth transformers trained as such establish depth recurrence in pretraining: latent recurrent-depth language models scale test-time compute by iterating a block \citep{geiping2025}, and looped-transformer analyses characterize implicit composition, depth extrapolation, and overthinking \citep{kohli2026}. Most directly, McLeish et al. convert pretrained non-recurrent language models to recurrent execution with a recurrence curriculum \citep{mcleish2025}, the closest precedent to the present setting. The present work differs in the split-bridge re-entry design, the exact intermediate-state supervision objective, and the demonstration that the retrofit also installs at an adapter budget over frozen base weights (Section 6.2). The system here uses explicit recurrent iterations rather than a root solver, and sequence-level probabilistic halting rather than per-token halting. Its main distinction is the retrofit setting: a pretrained causal language model is partitioned and looped after pretraining, making identity preservation and re-entry distribution alignment central experimental concerns.

\hypertarget{retrofitting-pretrained-transformers}{%
\subsection{Retrofitting pretrained transformers}\label{retrofitting-pretrained-transformers}}

Kasai et al. convert pretrained transformers into recurrent alternatives through a swap-then-finetune procedure targeting efficient autoregressive attention \citep{kasai2021}. LoRA freezes pretrained weights and adds low-rank trainable matrices \citep{hu2021}. The surgery here differs in purpose and location: the original attention implementation remains, while a middle transformer region is reused over computation depth. The structural surgery is identical at both of this paper\textquotesingle s training budgets. What varies is the adaptation inside it: full-block unfreezing of the looped region in one configuration, and rank-16 LoRA over the looped projections plus the bridge in the other. The adapter deltas are themselves weight-tied across loop iterations, so the low-rank correction is applied at every depth and the adapter becomes part of the recurrent operator. The adapter configuration thus uses low-rank adaptation not in its usual role of steering behavior within a fixed architecture, but to install a new computational mechanism through one. Prior work has converted pretrained transformers to recursively shared blocks with layer-wise low-rank adapters \citep{bae2024}, and the configuration here differs in sharing one rank-16 adapter across repeated executions of the surgically isolated block, jointly with the split bridge and the intermediate-state curriculum. Qwen2.5 provides the pretrained dense substrate \citep{qwen2024}.

\hypertarget{latent-iterative-and-stochastic-computation}{%
\subsection{Latent iterative and stochastic computation}\label{latent-iterative-and-stochastic-computation}}

The token-space alternative, writing intermediate steps out as text, descends from scratchpad training and chain-of-thought prompting \citep{nye2021,wei2022}, and serves here as the serialized dense control (Section 9). A parallel family reasons in continuous latent space along the sequence axis: Coconut feeds final hidden states back as input embeddings under a curriculum that progressively replaces chain-of-thought tokens \citep{hao2024}, with successors distilling explicit chains into continuous ones by self-distillation \citep{shen2025}, compressing dense latent traces \citep{cheng2024}, and theory establishing, in a graph-reachability construction, that continuous thoughts can encode multiple search frontiers in superposition \citep{zhu2025}. The recurrence here differs on four measured axes: vertical weight-tied depth rather than horizontal state feedback, installed per-step latent supervision rather than answer-only curricula, forced and separately studied depth selection rather than fixed thought budgets, and a characterized support-dependent extrapolation frontier. Where that theory reads one latent as holding many candidates in superposition, the counterfactual splice measurements here read the model\textquotesingle s state as one program rolled forward. The families are complementary.

Recent recurrent-depth studies also investigate implicit multi-hop composition and depth extrapolation, including the risk that excessive loops degrade predictions \citep{kohli2026}. GRAM, extending the HRM and TRM line of small from-scratch recursive reasoners \citep{wang2025,jolicoeur2025}, models recursive reasoning as a distribution over stochastic latent trajectories and trains a conditional prior against a target-conditioned posterior \citep{baek2026}. This manuscript evaluates the deterministic substrate study only.

\hypertarget{architecture}{%
\section{Architecture}\label{architecture}}

This section specifies the surgery: the three-region split, the re-entry bridge that closes the loop, what the loop does and does not change, halting, and the parameter and compute accounting.

\hypertarget{prelude-recurrent-block-and-coda}{%
\subsection{Prelude, Recurrent Block, and Coda}\label{prelude-recurrent-block-and-coda}}

Let the pretrained transformer contain ordered layers split at indices \texttt{(a,\ b)}:

\begin{verbatim}
tokens -> Prelude[0:a] -> Recurrent Block[a:b] -> Coda[b:L] -> LM head
\end{verbatim}

For Qwen2.5-0.5B-Instruct: 24 decoder layers (hidden width 896; 14 query heads with grouped-query attention over 2 key-value heads, head dimension 64; SwiGLU MLPs of intermediate width 4864; RoPE; RMSNorm; tied embeddings), split at a = 6 and b = 18. The Prelude computes an input-grounded representation \texttt{p}. The recurrent state begins from that representation and is updated by the shared middle block. After the selected number of loops, the Coda maps the final state into the pretrained output space (Figure 1).

For one loop, the wrapper executes each pretrained layer exactly once in its original order. Attention masks, position identifiers, normalization, dtype, and causal semantics are preserved. Recurrent additions are inactive on this identity route, which is the design constraint that makes loop-1 general-capability evaluation a measure of preservation rather than of a different model.

\begin{figure}[!tb]
\centering
\includegraphics[width=\linewidth]{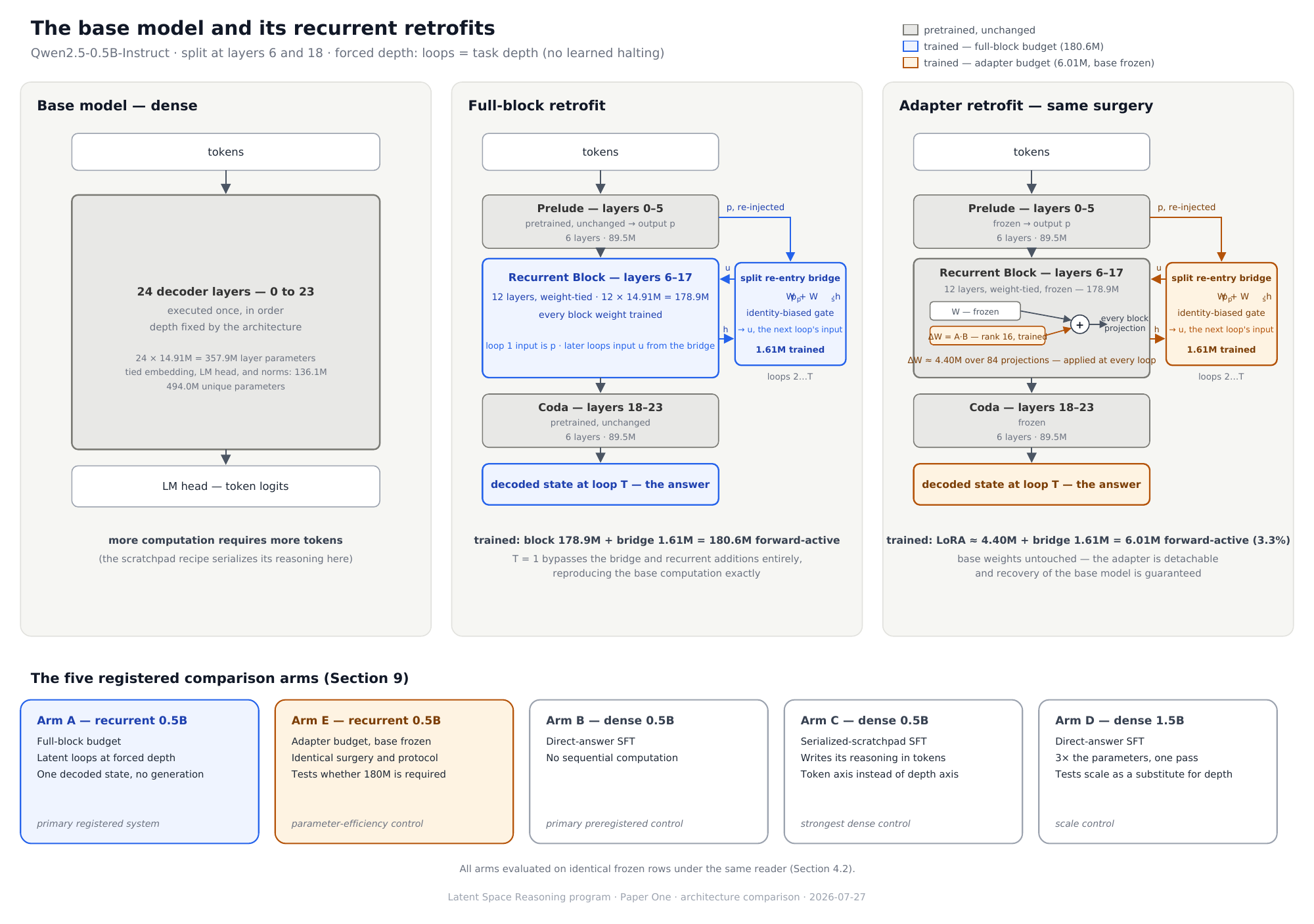}
\caption{The base model and the two retrofit budgets over one surgery. Left: the dense Qwen2.5-0.5B base, 24 layers executed once, where additional computation requires additional tokens. Center: the full-block retrofit, with the 12-layer weight-tied Recurrent Block and the split re-entry bridge trained (178.9M plus 1.61M, 180.6M forward-active) and the Prelude and Coda unchanged. Right: the adapter retrofit over the identical surgery, with every pretrained weight frozen and rank-16 LoRA on the block projections plus the bridge trained (6.01M forward-active). In both retrofits the bridge combines the carried state with the re-injected Prelude output under an identity-biased gate on loops 2 through T, and T = 1 bypasses the recurrent additions and reproduces the base computation exactly. The bottom strip maps the five registered comparison arms of Section 9 onto these architectures.}
\label{fig:arch}
\end{figure}

\hypertarget{the-re-entry-bridge-and-the-corrected-loop-closure}{%
\subsection{The re-entry bridge and the corrected loop closure}\label{the-re-entry-bridge-and-the-corrected-loop-closure}}

The later-loop update is input-injected recurrence. Writing p for the Prelude output and h\_t for the block output of iteration t, both of shape (B, S, 896):

\begin{verbatim}
u_t     = Reentry(p, h_t)
h_(t+1) = RecurrentBlock(u_t)
\end{verbatim}

The implemented split bridge learns separate projections of the persistent state and the Prelude representation, W\_p, W\_s in R\^{}\{896x896\}, recombined with an identity-biased gate and optional re-entry normalization (Figure 2). The split into separate parameters is what permits a distinct learning-rate group on the Prelude projection. The Prelude contribution is applied only on re-entry, so the one-loop identity path is unchanged. The re-injection is essential: without fresh access to the Prelude representation, iterated application drives states off the pretraining manifold within a few loops, so the bridge\textquotesingle s job is precisely to keep each iteration grounded in the input while the carried state accumulates computation. The implementation is verified by static graph inspection, per-loop gradient matrices, finite-difference checks, covariance and norm diagnostics, and forced-loop artifact checks. Listing~\ref{lst:bridge} shows the forward path.

\begin{lstlisting}[float=tb,caption={The split-bridge forward (\texttt{models/bridge.py}).},label=lst:bridge]
prelude = self.prelude_norm(prelude)
if self.split_projection:
    translated = self.prelude_proj(prelude) + self.state_proj(work)
else:
    translated = self.proj(torch.cat([prelude, work], dim=-1))
gate = self.bridge_gate.to(device=hidden_states.device, dtype=work.dtype)
return (work + gate * (translated - work)).to(dtype=input_dtype)
\end{lstlisting}

\begin{figure}[!tb]
\centering
\includegraphics[width=\linewidth]{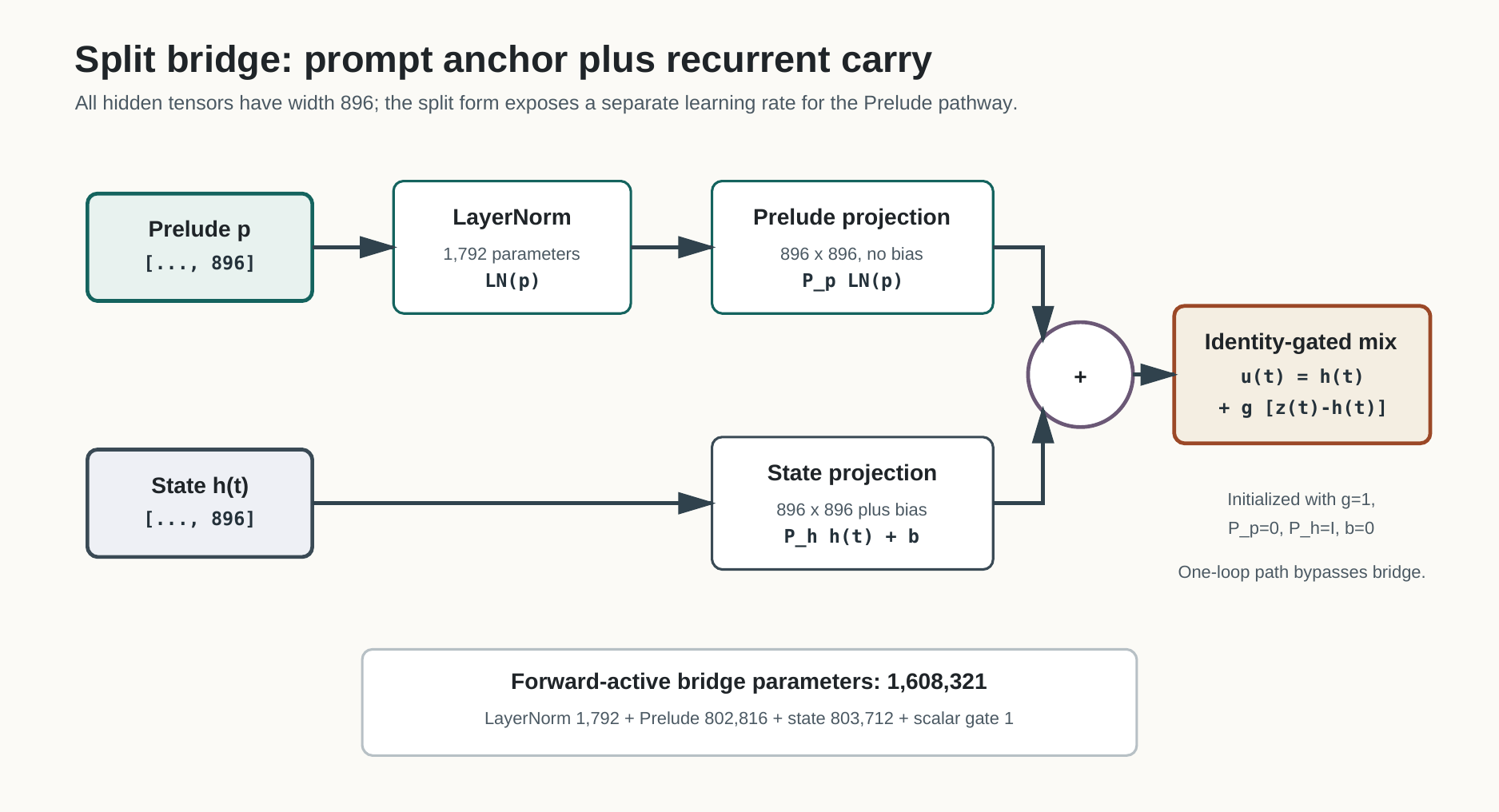}
\caption{The split bridge in tensor detail. Separate learned projections of the re-injected Prelude output p and the carried state $h_t$, combined under the identity-biased gate and normalization. The bridge is the sole cross-iteration channel.}
\label{fig:bridge}
\end{figure}

\hypertarget{what-the-loop-is-and-is-not}{%
\subsection{What the loop is and is not}\label{what-the-loop-is-and-is-not}}

The loop replaces \emph{depth}, not sequence. Within one iteration, the block runs standard causal self-attention over the S token positions. Across iterations, no attention connects loop t to loop t-1: the only channel by which iteration t sees iteration t-1 is the bridged state. There is no KV-cache growth with T, no context growth, and the sequence length is constant. Each iteration is one full-width parallel transformer pass. Serial depth accumulates in T, not in tokens. A recorded evaluation artifact fixes the shapes: stacked per-loop logits at T = 14 over a 100-token sequence are (1, 1, 14, 100, 151936).

\hypertarget{halting-and-forced-depth-evaluation}{%
\subsection{Halting and forced-depth evaluation}\label{halting-and-forced-depth-evaluation}}

A sequence-level halting head produces a distribution over loop counts. The training objective can combine task loss, intermediate-state losses, and a regularizer toward a centered geometric prior. Mechanism experiments use forced loop counts throughout this paper, separating two questions: can the block execute a useful transition at depth t, and can a router select the appropriate depth. This paper resolves the first question. The second is deliberately deferred: every result below uses forced depths so the mechanism is measured independently of any selector, and per-loop emissions are fully evaluated at every iteration. The learned selector, when finally tested alone over the working mechanism, produced a registered bounded negative (Section 12). Depth discovery, as opposed to depth reading, is left to future work with a redesigned controller.

\hypertarget{parameter-and-compute-accounting}{%
\subsection{Parameter and compute accounting}\label{parameter-and-compute-accounting}}

Per decoder layer, 14.91M parameters (attention 1.84M under grouped-query attention, MLP 13.07M). The historically optimizer-marked trainable total was 182,163,457: the 12-layer block at 178,948,608 plus a bridge module whose optimizer-marked 3,214,849 parameters include a legacy 1792-to-896 concatenation projection (1,606,528 parameters) that split-mode forward execution bypasses entirely, so it receives no functional gradient. Current code excludes the bypassed projection from optimizer groups, so new runs mark 180,556,929. The forward-active trained count is therefore 180,556,929: the block plus 1,608,321 active bridge parameters (prelude LayerNorm, two 896x896 projections, one with bias, and the scalar identity-biased gate). Both numbers are reported, the first for lineage accuracy, the second as the functional budget. A retired stability damper remains in the code at strength 0.0 (Appendix A.4). By either count, the full-block budget is not parameter-efficient fine-tuning. The parameter-efficient configuration referenced throughout as the adapter budget is rank-16 LoRA over all block projections plus this bridge, 6.01M forward-active parameters (Section 6.2). Three numbers are reported throughout. The forward-active model contains 495,641,089 unique parameters (496M rounded), counting the tied embedding and LM-head weight once. The full wrapper object holds 498,169,963 unique parameters, since it retains dormant compatibility and auxiliary tensors, and its raw state dict exposes 634,305,515 tensor elements because the tied weight appears under two keys. Active parameters per pass at T = 1 are the full base stack. Active compute at T loops equals the Prelude and Coda once plus the block T times: the Prelude and Coda together span 12 of the 24 layers and the recurrent block the other 12, so T loops cost (1 + T)/2 of a base forward pass, or 1.00x, 2.50x, 4.50x, and 6.50x base-model FLOPs per token at T = 1, 4, 8, and 12. This paper does not assess relative FLOPs between recurrent and token-space scratchpads. A preregistered wall-clock measurement is reported in Section 9.5, and the areas of interest are capability at recurrent depth with zero context growth and parallel-width steps.

\hypertarget{training-and-measurement-protocol}{%
\section{Training and Measurement Protocol}\label{training-and-measurement-protocol}}

This section fixes the apparatus shared by every experiment: the supervision scheme, the reader protocol, the lineage rules, and the limits of cross-arm comparison.

\hypertarget{exact-intermediate-state-supervision}{%
\subsection{Exact intermediate-state supervision}\label{exact-intermediate-state-supervision}}

The recurrent training task is clearest through a worked example. A row presents a scrambled rule table, "Ben passes to Max. Max passes to Eve. Eve passes to Kai. ...", and asks: start at Ben and apply the rule three times --- who do you land on? The correct computation is stepwise, Ben to Max to Eve to Kai, and the architecture is designed so that each execution of the recurrent block performs exactly one step: after loop 1 the model\textquotesingle s state should encode Max, after loop 2 Eve, after loop 3 Kai, the answer. Because the state at every loop can be decoded with the same reader that reads final answers, each loop can be asked, in effect, "if you had to answer right now, who do you land on?", and graded.

Intermediate-state supervision is exactly that grading: for loop t, the decoded state receives cross-entropy against the registered intermediate target, loop 1 against Max, loop 2 against Eve, and so on. Outcome-only rows grade only the final state. Staged training first installs the transition with dense intermediate labels, then removes that scaffold, grading only final answers, to test whether the stepwise mechanism persists when nothing in the loss rewards the steps (Section 5.2).

\hypertarget{frozen-same-reader-evaluation}{%
\subsection{Frozen same-reader evaluation}\label{frozen-same-reader-evaluation}}

Evaluation in a looped model has a structural requirement: the model must be correct at every iteration against the known oracle chain, not merely at the end, so the reader that decodes each iteration is part of the experimental apparatus and must itself be validated. Two design rules follow. First, one reader everywhere: the same-reader protocol holds the prompt surface, output symbol vocabulary, row IDs, and decoder constant across loop counts and systems, so intermediate and final correctness are measured on the surface the model was trained to produce. Second, a reader is calibrated against a known oracle before its verdicts are trusted: a simulated perfect executor, which by construction contains the correct intermediate at every loop, must score perfectly under it. Concretely, for the depth-3 row of Section 4.1 the reader must ask, at loop 1, "is the state Max?", at loop 2, "is it Eve?", and only at loop 3, "is it Kai?", grading each iteration against its own correct intermediate rather than against the terminal answer. The reported reader passes this calibration exactly, and under it readout is lossless: on the primary frozen family, final-symbol accuracy equals internal chain accuracy at the final loop identically, 1,506/1,792 to the count, so "final-answer capability" and "chain capability" are one measurement.

Listing~\ref{lst:eval} shows the forced-read evaluation path.

\begin{lstlisting}[float=tb,caption={One forward pass, forced read at every loop (\texttt{eval/eval\_synthetic\_depth\_active\_labels.py}).},label=lst:eval]
with torch.no_grad():
    output = wrapper(
        input_ids=encoded["input_ids"],
        attention_mask=encoded["attention_mask"],
        max_loops=max(loop_counts),
        use_cache=False,
        return_dict=True,
        return_loop_logits=True,
    )
for loop in loop_counts:
    logits = select_forced_loop_logits(output, loop)
\end{lstlisting}

The primary synthetic family has 128 rows at each depth from 1 through 14, 1,792 rows total. The Phase A arms, the five systems of the registered dense comparison (Section 9), use identical row IDs. Paired exact sign/McNemar tests compare correctness row by row. A count-based Fisher gate was preregistered for the primary comparison.

\hypertarget{lineage-and-guardrails}{%
\subsection{Lineage and guardrails}\label{lineage-and-guardrails}}

Every promoted checkpoint has a SHA-256 evidence record. Continuations resolve both the exact source checkpoint and its source metrics before loading the model. A continuation cannot start below a registered retention floor. Full-block exploratory runs are marked \texttt{disposable\_measurement}, cannot produce successor lineage, and must delete non-promotable checkpoints after their measurement completes.

The same discipline extends to the paper\textquotesingle s claims. Every claim links to a durable evidence record in a machine-readable ledger, unsupported natural-benchmark and stochastic-width claims are excluded, and a machine-checked prohibited-phrases list in the repository\textquotesingle s test suite enforces the exclusions.

\hypertarget{comparison-limits}{%
\subsection{Comparison limits}\label{comparison-limits}}

The recurrent and dense arms share the model family, frozen rows, and output reader, but not identical training histories, token counts, optimizer trajectories, FLOPs, latency, or inference compute. Phase A is consequently a registered system comparison, not an architecture-only causal estimate.

\hypertarget{optimizer-interaction-and-mechanism-persistence}{%
\section{Optimizer Interaction and Mechanism Persistence}\label{optimizer-interaction-and-mechanism-persistence}}

This section reports two properties of the mechanism\textquotesingle s training: an optimizer interaction that silently defeats slice-level learning-rate control, and the persistence of the stepwise mechanism after its training scaffold is removed.

\hypertarget{learning-rate-control-in-combined-projections-an-optimizer-interaction}{%
\subsection{Learning-rate control in combined projections: an optimizer interaction}\label{learning-rate-control-in-combined-projections-an-optimizer-interaction}}

A recurrent retrofit mixes new parameters with pretrained ones, so differential learning rates are a design requirement. Here, the bridge\textquotesingle s Prelude projection required a raised rate, and the first implementation expressed it as gradient-slice scaling inside the combined re-entry projection, a single weight matrix whose input columns served both the re-injected Prelude representation and the carried state. That implementation does not deliver the intended larger realized update under either optimizer family, and the failure follows directly from the update rules, derived in Appendix E. Under Adam-family updates, per-element moment normalization cancels a constant gradient scale exactly, up to the epsilon floor. Under Muon, whole-matrix orthogonalization discards a global scale, and a slice factor perturbs the update\textquotesingle s direction without raising the slice\textquotesingle s effective rate. A per-group learning rate enters outside both normalizations and scales the realized update linearly, but ordinary parameter groups address whole tensors, so a slice inside one combined projection cannot receive its own rate without splitting the tensor.

Both predictions were confirmed empirically. Under AdamW, a controlled slice-level comparison measured the realized difference between scaled and unscaled slices at 1.3e-9, the epsilon-floor residue. Under Muon, the evidence is symptomatic rather than controlled: a development run with a 10x multiplier moved the Prelude weight only 7.36e-4 RMS, consistent with orthogonalization discarding the scale. A gradient-path audit ruled out connectivity as the cause. Gradients reached the bridge throughout, the loop-2 Prelude gradient median was 1.30e-2, the zero-gradient fraction was 0 at active loops, and a finite-difference check confirmed the path. The production design therefore splits the projection into independent Prelude and state parameters, places the Prelude projection in a dedicated AdamW parameter group carrying a true 10x rate with a startup assertion that the groups exist, and rejects rate multipliers under Muon outright. The group\textquotesingle s activity is verified at the realized-update level, with Prelude movement at 1.82e-3 RMS. Listing~\ref{lst:group} shows the group construction.

\begin{lstlisting}[float=tb,caption={The dedicated parameter group (\texttt{training/train\_unfrozen\_recurrent.py}).},label=lst:group]
if optimizer_name == "adamw" and prelude_params:
    rest = [p for p in params if id(p) not in prelude_param_ids]
    return torch.optim.AdamW([
        {"params": rest, "lr": lr, "weight_decay": weight_decay},
        {"params": prelude_params,
         "lr": prelude_lr_multiplier * lr,
         "weight_decay": cfg_float(cfg, "bridge_prelude_weight_decay", 0.0)},
    ])
\end{lstlisting}

Thus one important finding of this research is that when retrofitting a recurrent bridge into a pretrained transformer, any parameter that needs its own learning rate must be a separate tensor in its own optimizer group. Scaling a slice of a shared matrix\textquotesingle s gradient fails silently. Training proceeds, losses fall, and no error is raised, but the intended change never happens, under Adam because per-element normalization cancels it and under Muon because orthogonalization discards it. The failure is invisible in every standard training signal, so proof that a raised rate is operating requires a startup assertion on the parameter group and its learning rate, followed by a realized parameter-movement check, not the code that requests it.

\hypertarget{persistent-chain-after-scaffold-removal}{%
\subsection{Persistent chain after scaffold removal}\label{persistent-chain-after-scaffold-removal}}

This experiment tests whether the stepwise mechanism survives once the intermediate-label scaffold is removed. After staged chain training, the final 1,000 steps used outcome supervision only. The post-anneal checkpoint retained:

\begin{table}[!tb]
\centering
\caption{Post-anneal retention of the stepwise mechanism, full-block lineage.}
\label{tab:anneal}
\small
\begin{tabular}{lr}
\toprule
Measure & Result \\
\midrule
Active-label diagonal, depths 1-4 & \path{625/640\ =\ 97.7\%} \\
Above-diagonal states that continued iterating & \path{357/384\ =\ 93.0\%} \\
Above-diagonal states that held & \path{1/384\ =\ 0.3\%} \\
\bottomrule
\end{tabular}
\end{table}

The first row reflects decoding the intermediate states on held-out rows, after a final training phase which no longer graded them: 625 of 640 still held the correct intermediate answer, showing that the model solved problems step by step even when only answers were rewarded. Stepwise computation is how it works, not a performance for the grader. The second and third rows are the stricter check: force a depth-3 problem to run loops 4 through 8 and ask what appears in the extra states. A model that had learned "produce the answer and stop" would park, its state repeating the answer. A shortcut model would show unrelated states. Instead, 93.0\% of extra-loop states contained the correct continuation of the chain, the symbol reached by applying the rule once more, and the model essentially never parked (1 of 384). Producing the correct continuation is what one further application of the rule yields, so the two measurements rule out terminal-answer lookup. What was installed in the block is the operation "apply the rule once": running the block is applying the rule, and a depth-d answer is d applications of the rule. The practical corollary: because the model keeps iterating, running too many loops overshoots the answer, which is why depth selection is a separate problem (Section 3.4) rather than a free byproduct of the mechanism. The signature replicates at the adapter budget, and more strongly (Section 9.4).

\hypertarget{the-recurrent-mechanism-under-full-block-and-lora-fine-tuning}{%
\section{The Recurrent Mechanism Under Full-Block and LoRA Fine-Tuning}\label{the-recurrent-mechanism-under-full-block-and-lora-fine-tuning}}

The mechanism was installed and examined under two fine-tuning regimes over the identical surgery: full-block adaptation of the looped region, and rank-16 LoRA with every pretrained weight frozen. This section reports the two installations side by side, and the characterization that follows draws on both.

\hypertarget{full-block-fine-tuning}{%
\subsection{Full-block fine-tuning}\label{full-block-fine-tuning}}

Unfreezing the 12-layer recurrent block and training it with the bridge, 180.6M forward-active parameters (Section 3.5), installed the mechanism under the reference recipe: active-label accuracy of 1.000, 0.984, 0.984, and 0.922 at depths 1 through 4 after 6,000 steps. This lineage carries the persistence result of Section 5.2, the scaling characterization of Section 8, the verbal transfer results of Section 10, and the preservation battery of Section 7.

\hypertarget{lora-fine-tuning}{%
\subsection{LoRA fine-tuning}\label{lora-fine-tuning}}

The same mechanism installs with the pretrained base weights untouched: rank-16 LoRA over all block projections plus the bridge, 6,007,425 forward-active trainable parameters (7,613,953 optimizer-marked: the difference is the bypassed legacy projection of Section 3.5, and the accounting reconciles exactly with the matched build of Section 9.4). The configuration passed the depth-1-through-4 installation gate, first crossing at 4,000 cumulative steps and finishing at 64/64, 64/64, 60/64, and 53/64, with the base-weight hash unchanged at run end, the one-loop identity difference exactly 0.0, and capability passing the Tier-1 arithmetic check, a prespecified 64-row benchmark used as a capability guardrail throughout the program (60/64 before, 61/64 after). The adapter is detachable, so recovery of the base model is guaranteed by construction. Per-depth differences against the full-block reference are not statistically significant at 64 rows per depth (depth 4: 53 versus 59, one-sided Fisher p = 0.090), though the comparison is underpowered to claim parity.

\hypertarget{the-budget-comparison}{%
\subsection{The budget comparison}\label{the-budget-comparison}}

\begin{table}[!tb]
\centering
\caption{The two training budgets over the identical surgery.}
\label{tab:budgets}
\small
\resizebox{\linewidth}{!}{%
\begin{tabular}{lrrll}
\toprule
Budget & Forward-active & Steps to gate & D1-D4 at finish & Preservation mode \\
\midrule
Full-block & 180.6M & 6,000 & 64, 63, 63, 59 (of 64) & battery non-inferiority (Section 7) \\
Adapter (rank 16) & 6.01M & 4,000 & 64, 64, 60, 53 (of 64) & base weights untouched, by construction \\
\bottomrule
\end{tabular}
}
\end{table}

The adapter budget is 3.3 percent of the full-block budget. When both versions are trained with the same deep curriculum, the comparison shows a crossover that depends on depth (Section 9.4). Overall accuracy is statistically indistinguishable. The LoRA version is slightly better on the depths it was trained on and for the first three depths beyond them, through depth 11. The full-block version is clearly better at depths 12 through 14. Installing the mechanism and extrapolating far past training therefore have different capacity requirements. Three further results characterize the LoRA lineage. Its mechanism persists strongly after the intermediate-label scaffold is removed. Without the verbal training the full-block lineage received, it shows minimal zero-shot transfer to the verbal task surfaces, though its installed history accelerates matched verbal training (Section 10.1). And its retention probe was stopped by the pre-specified guardrails, with the acquisition-retention boundary holding at this budget too (Sections 9.4, 10, and 11.4).

\hypertarget{general-capability-preservation-under-installation}{%
\section{General-Capability Preservation Under Installation}\label{general-capability-preservation-under-installation}}

Installing the mechanism retrained the full 12-layer block, so preservation of the base model\textquotesingle s general capability cannot be assumed. The identity route of Section 3.1 guarantees base behavior only while the block weights are unchanged, and training changed them. Whether loop-1 behavior survived is an empirical question, and a preregistered battery answered it. ARC-Easy and ARC-Challenge were evaluated in full (5,197 and 2,590 paired rows per checkpoint) at loop 1, on four promoted checkpoints spanning the keeper lineage, the sequence of frozen reference checkpoints the program promotes, against the shared base model. The noninferiority margin was three points of accuracy, with exact paired tests Bonferroni-corrected across the eight checkpoint-by-benchmark comparisons, alongside a permutation zero-shot control and the prespecified natural-surface check.

Every promoted checkpoint was non-inferior, and all eight comparisons passed the margin. The single nominal dip, the N24 checkpoint (the deepest-trained configuration, Section 8.1) on ARC-Easy at 14 rows (3,962 versus 3,976 of 5,197, unadjusted p = 0.0385), did not survive correction (corrected p = 0.308). The permutation control on fresh tables scored 9,841/9,984 (98.6\%), with per-depth parity within the 0.05 tolerance.

The restriction to loop 1 is not cosmetic. A forced-depth diagnostic on an earlier unfreeze-era checkpoint matched the base on ARC-Challenge at loop 1 (89 versus 88 content-scored, 155 versus 155 permutation-scored) but degraded progressively at deeper forced loops, decisively so by loop 8 (71 versus 155, p = 3.8e-14). Off task, extra loops are not free. The trained model preserves base capability where it is designed to be evaluated, at loop 1, and solving the depth-selection problem of Section 3.4 is what would make deeper allocation safe.

The preservation claim is this: converting 12 of 24 layers into a trained recurrent block, a 182M-parameter adaptation, did not measurably degrade the evaluated battery at any point along the installation lineage. The statement is scoped to the evaluated battery and does not assert universal preservation. It carries weight because the same instrument detects real damage in Sections 11.1 through 11.3, so the passing results here are informative rather than vacuous. The adapter budget is not part of this battery. It is preserved by construction, since its base weights never change (Section 6.2).

\hypertarget{support-depth-scaling}{%
\section{Support-Depth Scaling}\label{support-depth-scaling}}

The central scaling question is how far the installed operation extrapolates beyond the depths it was trained on, and how that reach grows with training support.

\hypertarget{the-cross-support-frontier}{%
\subsection{The cross-support frontier}\label{the-cross-support-frontier}}

Measuring reach requires a threshold which was fixed by simple probability at the task design, before any scaling experiment was run. Specifically, each step of the task is a four-option multiple-choice question that the model must answer correctly, so the chance outcome at each step is 0.25. The threshold is the per-step accuracy where a four-step chain succeeds as often as a single chance guess, the fourth root of one fourth, or approximately 0.71. Below that level, chaining four steps is no better than guessing once. The single guess is the relevant comparison because it is the strategy available without the mechanism: simply guessing the final answer directly. The threshold is a design heuristic under an independence approximation, not an inferential cutoff, and it was held constant across every configuration compared. A registered sensitivity check repeated the analysis at thresholds 0.60 and 0.80, and the spread of frontier-to-support ratios across supports stayed below 0.09 at all three thresholds, so the near-constant scaling reflects the sharpness of the accuracy decline at the frontier rather than the placement of the threshold.

Four training configurations were run: supports 4, 6, and 8 on the 16-symbol alphabet (N16) and support 12 on the 24-symbol alphabet (N24), each evaluated on held-out rows at every depth, with the frontier defined as the depth where active-label accuracy crosses the threshold (linear interpolation between adjacent depths):

\begin{table}[!tb]
\centering
\caption{Threshold-crossing frontier by supervised support at consolidation.}
\label{tab:frontier}
\small
\begin{tabular}{rlrr}
\toprule
Support & Alphabet & Frontier & Frontier / Support \\
\midrule
4 & N16 & \texttt{5.75} & \texttt{1.44} \\
6 & N16 & \texttt{9.01} & \texttt{1.50} \\
8 (consolidated) & N16 & \texttt{11.61} & \texttt{1.45} \\
12 & N24 & \texttt{17.93} & \texttt{1.49} \\
\bottomrule
\end{tabular}
\end{table}

The frontier tracked support at a near-constant ratio across a threefold change in support and a change of alphabet (Figure 3). Teaching the operation to depth d buys reliable execution to roughly 1.5d, and the surplus is proportional to the training, not fixed, but with two qualifications. First, training dose must be scaled with support: at the support-6 step budget, the support-8 frontier reached only 9.75, and 2,000 additional steps restored the 1.45 ratio, so fixed-dose comparisons understate larger supports. Second, consolidation speed varies materially by seed: support-6 outcomes at matched steps spanned 6.95 to 9.01, while under the dose protocol the replicates finished at 9.50 and 10.00, inside the registered band. The scaling pattern repeats across seeds, provided the training dose is sufficient. The N24 point was only run with a single training seed. The ratio also survived a thirtyfold budget cut, the LoRA arm producing a nearly identical support-8 frontier (Section 9.4). The ratio does not extend indefinitely, and the ceiling that ends it is measured next.

\begin{figure}[!tb]
\centering
\includegraphics[width=\linewidth]{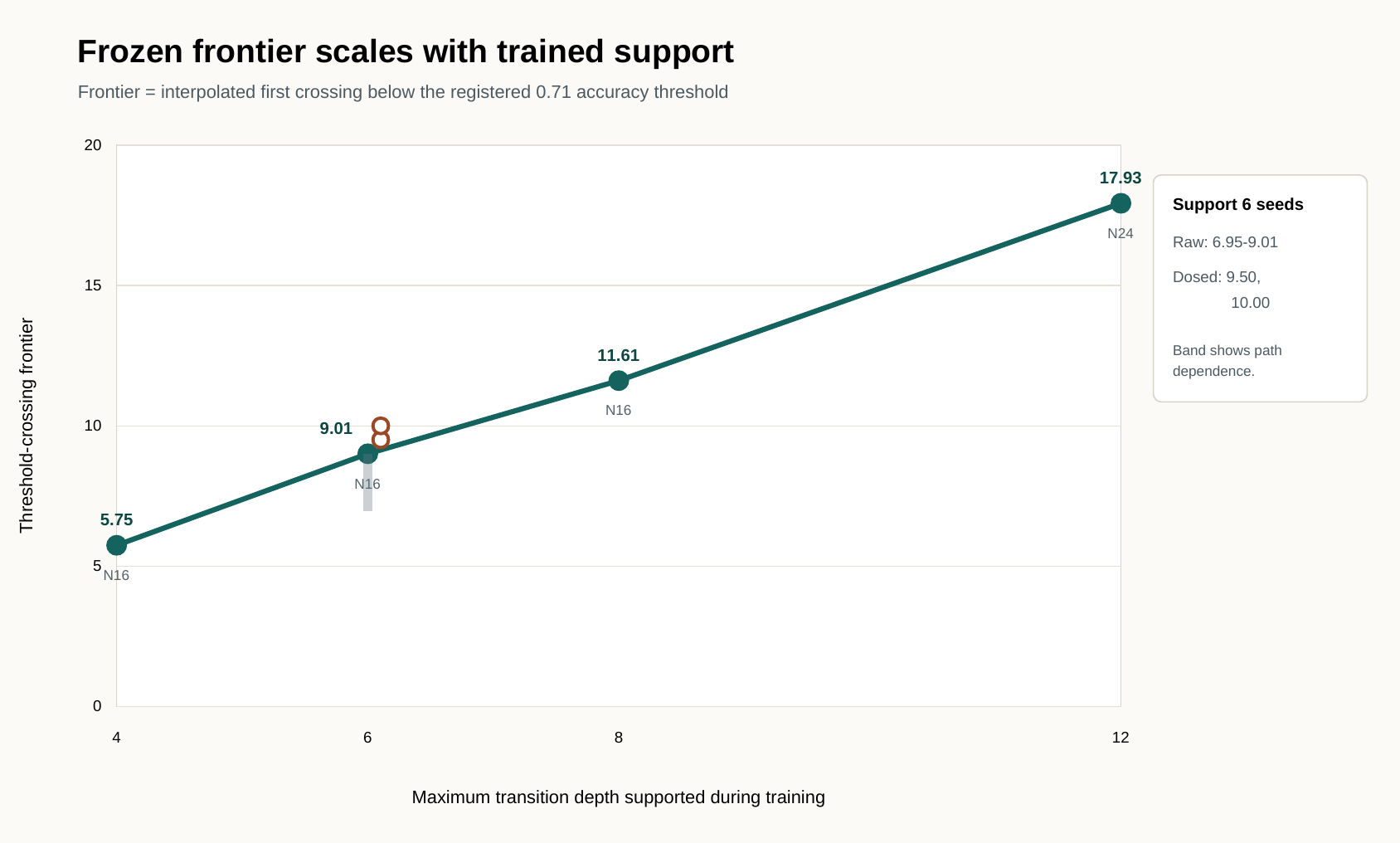}
\caption{Threshold-crossing frontier versus supervised support at consolidation, four supports across two alphabet sizes. The support-6 point shows the three-seed band.}
\label{fig:frontier}
\end{figure}

\hypertarget{the-n24-frontier-and-its-ceiling}{%
\subsection{The N24 frontier and its ceiling}\label{the-n24-frontier-and-its-ceiling}}

The N24 program evaluated every depth from 1 through 22 on a frozen set. The final step-6,000 checkpoint produced:

\begin{table}[!tb]
\centering
\caption{N24 support-12 accuracy by depth, step-6{,}000 checkpoint.}
\label{tab:n24}
\small
\begin{tabular}{rr}
\toprule
Depth & Accuracy \\
\midrule
10 & \texttt{97.7\%} \\
12 & \texttt{97.7\%} \\
14 & \texttt{91.4\%} \\
16 & \texttt{85.9\%} \\
18 & \texttt{70.3\%} \\
20 & \texttt{46.1\%} \\
22 & \texttt{10.9\%} \\
\bottomrule
\end{tabular}
\end{table}

At step 2,000, depth-14 accuracy was 69.5\% and depth-18 accuracy was 10.2\%: depths that failed at the lower dose passed at the higher one, the same dose pattern as Section 8.1, here observed within a single run. Additional support moved the frontier substantially but did not create indefinite algorithmic extrapolation. The result is a support-dependent extrapolation frontier with a measurable tail ceiling.

\hypertarget{registered-comparison-against-dense-baselines}{%
\section{Registered Comparison Against Dense Baselines}\label{registered-comparison-against-dense-baselines}}

This section reports the preregistered comparison between the recurrent system and the practical dense alternatives. The design and its primary gate were fixed in advance rather than post hoc, and every arm is built from the same model family and evaluated on the same frozen rows under the same reader. Four controls isolate what the recurrence contributes (Figure 1, bottom). Arm B, a direct-answer 0.5B, tests whether the task can be learned with no sequential computation. Arm C, a 0.5B that writes its reasoning out in tokens, is the standard alternative to latent iteration. Arm D, a direct-answer 1.5B, tests whether additional parameters substitute for depth. Arm E repeats the recurrent system at the 6M adapter budget, testing whether the result requires the 180M training budget.

It should be noted that training lineage and inference compute are not matched across arms, so this is a registered system comparison rather than an architecture-only estimate (Section 4.4). Within that frame, the summary is simple. The scratchpad matches the recurrent arms inside its learned horizon but collapses beyond it, the recurrent arms retain accuracy past that horizon and win overall, and they are also the fastest under the interactive conditions measured in Section 9.5.

\hypertarget{systems}{%
\subsection{Systems}\label{systems}}

\begin{table}[!tb]
\centering
\caption{The five Phase A systems.}
\label{tab:systems}
\small
\resizebox{\linewidth}{!}{%
\begin{tabular}{lll}
\toprule
Arm & System & Recipe \\
\midrule
A & Recurrent Qwen2.5-0.5B & forced-depth same-reader recurrent system \\
B & Dense Qwen2.5-0.5B & direct-answer SFT \\
C & Dense Qwen2.5-0.5B & serialized-scratchpad SFT (chain of thought written out in tokens) \\
D & Dense Qwen2.5-1.5B & direct-answer scale control \\
E & Recurrent adapter-budget 0.5B & rank-16 LoRA plus bridge (6.0M forward-active) at the full-block arm\textquotesingle s training protocol \\
\bottomrule
\end{tabular}
}
\end{table}

\hypertarget{aggregate-and-tail-results}{%
\subsection{Aggregate and tail results}\label{aggregate-and-tail-results}}

The dense arms are scored by a reader that accepts a completed answer regardless of any text the model emits after it. The dense models were not trained to produce an end-of-sequence token and often generate a correct answer and then continue, so this is a generous reading of the dense arms. The recurrent arms read a single decoded state, so the question does not arise. Appendix C describes the reader in detail.

\begin{table}[!tb]
\centering
\caption{Phase A aggregate and tail results on the frozen 1{,}792-row family.}
\label{tab:phasea}
\small
\resizebox{\linewidth}{!}{%
\begin{tabular}{lrrrr}
\toprule
Arm & Correct / 1,792 & Accuracy & Depths 11-14 & Tail accuracy \\
\midrule
A, recurrent 0.5B (full block) & \texttt{1,506} & \texttt{84.04\%} & \path{272/512} & \texttt{53.13\%} \\
E, recurrent 0.5B (adapter, 6.0M) & \texttt{1,501} & \texttt{83.76\%} & \path{247/512} & \texttt{48.24\%} \\
B, dense 0.5B direct & \texttt{496} & \texttt{27.68\%} & \path{60/512} & \texttt{11.72\%} \\
C, dense 0.5B scratchpad & \texttt{1,292} & \texttt{72.10\%} & \path{13/512} & \texttt{2.54\%} \\
D, dense 1.5B direct & \texttt{656} & \texttt{36.61\%} & \path{58/512} & \texttt{11.33\%} \\
\bottomrule
\end{tabular}
}
\end{table}

The primary preregistered comparison was A versus B. A cleared the count-based gate at thirteen consecutive depths, 2 through 14, with depth 1 a 128/128 tie. A versus C was a labeled extension against the strongest dense control.

\begin{table}[!tb]
\centering
\caption{Row-paired comparisons against the primary and strongest dense controls.}
\label{tab:paired}
\small
\resizebox{\linewidth}{!}{%
\begin{tabular}{lrrrrr}
\toprule
Paired comparison & Helped & Hurt & Tied & Net & Two-sided exact p \\
\midrule
A vs B & \texttt{1,048} & \texttt{38} & \texttt{706} & \texttt{+1,010} & \texttt{5.72e-257} \\
A vs C & \texttt{262} & \texttt{48} & \texttt{1,482} & \texttt{+214} & \texttt{7.81e-37} \\
\bottomrule
\end{tabular}
}
\end{table}

The strongest evidence is the horizon crossover. The scratchpad ran at or near ceiling through depth 9 and held 127/128 at depth 10, then fell to 13/128 at depth 11 and to zero at depths 12 through 14, 2.5\% over the tail where the recurrent system retained 53.1\%. A plausible hypothesis, not a measured result, is that serialization length outruns what a 0.5B model holds coherent while the latent rollout keeps composing at constant context length.

\begin{figure}[!tb]
\centering
\includegraphics[width=\linewidth]{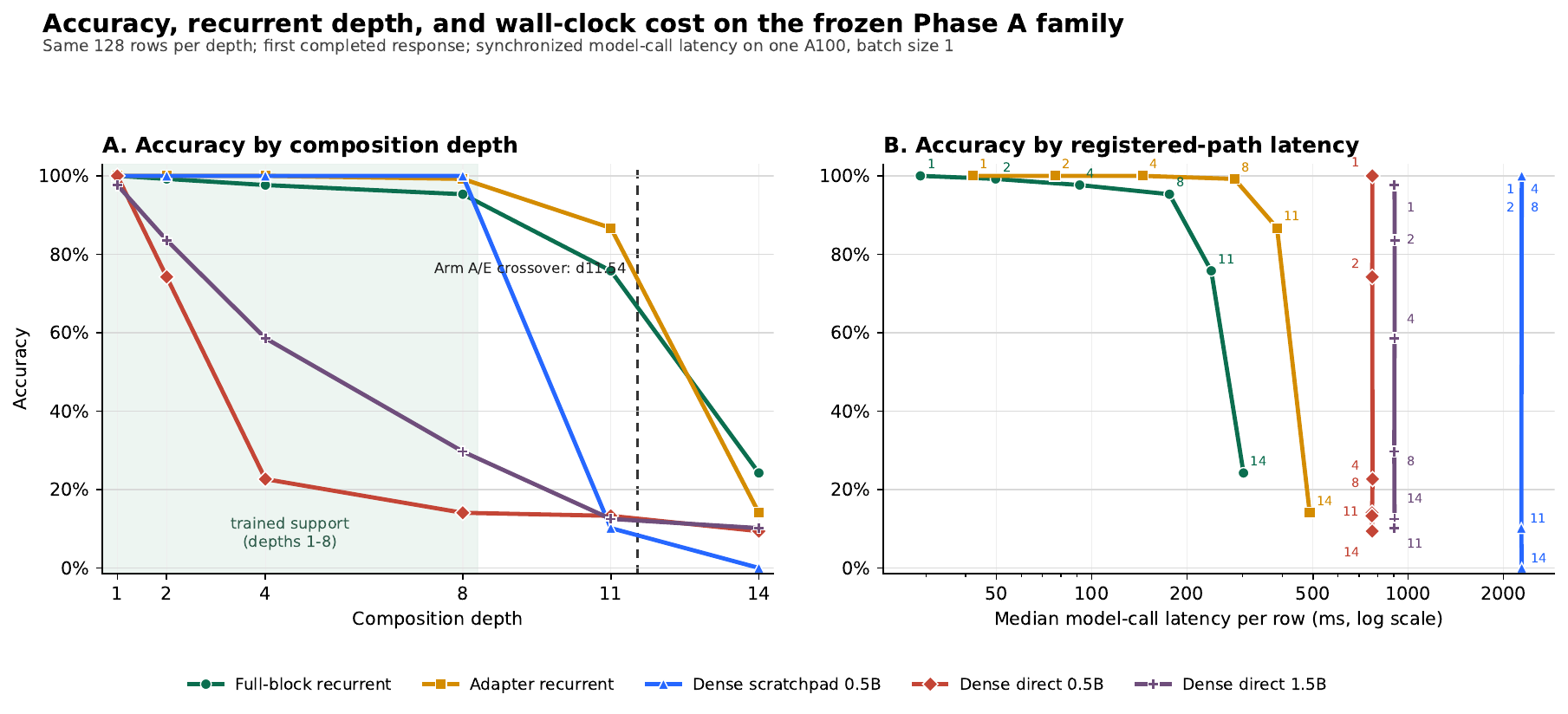}
\caption{Accuracy, recurrent depth, and wall-clock cost on the frozen Phase A family, 128 identical rows per depth. Panel A: accuracy by composition depth for all five arms, with trained support (depths 1 through 8) shaded and the Arm A/Arm E accuracy crossover marked at depth 11.54. Panel B: the same accuracies against median registered-path model-call latency (log scale). The recurrent arms trace the accuracy-latency frontier, while each dense arm pays its fixed generation cost at every depth. Values trace to the Phase A receipts and the synchronized latency receipt.}
\label{fig:phasea}
\end{figure}

\hypertarget{checkpoint-extension-results}{%
\subsection{Checkpoint-extension results}\label{checkpoint-extension-results}}

The dense arms were trained on from step 2,000 to step 4,000 to test whether additional training would change the comparison. The extension deltas were small and statistically unresolved: the direct dense arm added one net row (p = 1.000), the scratchpad arm added one (p = 1.000), and the 1.5B dense arm added 13 (p = 0.512). The totals place the 1.5B direct arm above the 0.5B direct arm, 656 versus 496, so the scale control runs in the direction an undertraining account would predict. That account remains an interpretation rather than a finding: the 1.5B arm\textquotesingle s improvement was statistically unresolved, and neither direct arm established convergence.

\hypertarget{the-adapter-budget-arm-a-depth-crossover}{%
\subsection{The adapter-budget arm: a depth crossover}\label{the-adapter-budget-arm-a-depth-crossover}}

The adapter arm (Arm E) trained the rank-16 configuration at the full-block arm\textquotesingle s exact staged protocol, 10,500 matched steps from a fresh base surgery with every pretrained parameter frozen, and was evaluated on the identical frozen rows with exact pairing. On pooled accuracy the two budgets were nearly identical: 1,501 versus 1,506 correct of 1,792, well within the registered 3-point margin (paired two-sided p = 0.813, with 140 rows helped, 145 hurt, and 1,507 tied). The comparison nevertheless failed its registered gate, because the gate also required the adapter arm to stay within 8 rows of the full-block arm at every individual depth, and it fell short at depths 12, 13, and 14, by 12, 14, and 13 rows.

Interestingly, on trained support, depths 1 through 8, the adapter arm was better, 1,021/1,024 versus 1,005/1,024 (post-hoc grouped paired p = 0.000855). On near extrapolation, depths 9 through 11, it stayed ahead, 344/384 versus 326/384 (post-hoc p = 0.0693), including a 14-row advantage at depth 11 (nominal unadjusted p = 0.0436, reported descriptively across 14 depths). Only at depths 12 through 14 did full-block training win, 175/384 versus 136/384 (post-hoc p = 0.00394). The interpolated accuracy crossover between the two arms falls at depth 11.54, the value marked in Figure 4. Note that the grouped p-values are post-hoc and were not subject to correction for multiple testing.

The 6M parameter adapter, just 3.3 percent of the full-block budget with the pretrained base untouched, can install the looping mechanism with a similar degree of depth extrapolation: its threshold-crossing frontier was 11.56 at support 8, against the full-block arm\textquotesingle s 11.61. The full-block budget buys a gentler decline beyond the frontier rather than a farther frontier, and the far-tail difference at depths 12 through 14 is not attributed among adapter rank, frozen-base geometry, bridge capacity, and optimization, since no rank sweep was run.

Persistence also holds at this budget, and slightly exceeds the full-block reference. After 1,000 outcome-only steps from the adapter arm\textquotesingle s final checkpoint, the active-label diagonal held at 636/640 (99.4\%, against the full-block reference of 625/640), 380/384 (99.0\%) of above-diagonal states continued iterating, and none held (references 357/384 and 1/384), the strong band of the registered battery. One supporting note: reaching the trained-support result required the matched extra depth-8 dose stage, where a mid-training check of 115/128 rose to 127/128 with continued matched training, ruling out an in-support adapter ceiling.

\hypertarget{wall-clock-latency}{%
\subsection{Wall-clock latency}\label{wall-clock-latency}}

A preregistered latency measurement covered all five systems on the frozen rows: one NVIDIA A100-SXM4-80GB, bfloat16 weights, batch size 1, greedy decoding, and each system\textquotesingle s registered evaluation path. The primary values are synchronized model-call medians per row (ms), with the median generated-token count:

\begin{table}[!tb]
\centering
\caption{Synchronized model-call latency medians per row (ms) with median generated-token counts.}
\label{tab:latency}
\small
\resizebox{\linewidth}{!}{%
\begin{tabular}{rrrrrr}
\toprule
Depth & Full-block recurrent & Adapter recurrent & Scratchpad & Direct 0.5B & Direct 1.5B \\
\midrule
1 & 28.8 & 42.1 & 2,287.9 & 770.3 & 905.1 \\
2 & 49.8 & 76.6 & 2,289.5 & 769.7 & 906.5 \\
4 & 91.7 & 145.0 & 2,288.2 & 771.6 & 905.5 \\
8 & 176.1 & 282.5 & 2,284.0 & 768.8 & 903.7 \\
11 & 238.7 & 385.2 & 2,276.9 & 768.2 & 905.6 \\
14 & 301.8 & 488.4 & 2,283.5 & 769.8 & 904.2 \\
Tokens (median) & 1 & 1 & 96 & 32 & 32 \\
\bottomrule
\end{tabular}
}
\end{table}

As might be expected, the recurrent arms scale linearly with depth at constant sequence length. Each additional loop costs approximately 21 ms for the full-block arm and 34 ms for the adapter arm, so a depth-1 answer takes 28.8 and 42.1 ms respectively, while a depth-14 answer takes 301.8 and 488.4 ms. The dense arms are flat across depth because their generation caps are fixed. The scratchpad path takes approximately 2.28 seconds to write its 96 tokens, and the direct systems take 0.77 and 0.90 seconds, whatever the problem depth. Although it might be expected, scratchpad latency did not grow with task depth. Under these interactive conditions, both recurrent systems answer faster than every dense system at every measured depth. At depth 14, the full-block arm is 7.6 times faster than the scratchpad. The comparison is descriptive, from one hardware configuration at batch size 1, and makes no claim about batched throughput. Appendix B reports the baseline-subtracted decomposition and its reference medians.

Panel B of Figure 4 plots the two measurements together: accuracy against median model-call latency at each measured depth. The recurrent arms trace the accuracy-latency frontier, while each dense arm pays its fixed generation cost at every depth.

\hypertarget{controlled-natural-surface-transfer}{%
\section{Controlled Natural-Surface Transfer}\label{controlled-natural-surface-transfer}}

The goal of this experiment was to evaluate whether the installed recurrent mechanism, trained entirely on symbolic rows under supervised fine-tuning, carries over to the same computation written in natural language, and so whether the latent reasoning loop can generalize beyond its synthetic installation task. The full-block natural keeper was evaluated on two generated verbal surfaces, scored with the same symbol reader as every other experiment. The keeper evaluated here received verbal training, and the zero-shot controls below separate what that training contributes. A relay row reads "Ben hands the token to Max. Max hands the token to Eve. Eve hands the token to Kai. Starting from Ben, who holds the token after three handoffs?" A pointer row renders the same structure as reference-following. "Ben\textquotesingle s contact is Max. Max\textquotesingle s contact is Eve. Eve\textquotesingle s contact is Kai. Following contacts three times from Ben, who is reached?" These examples are paraphrased for exposition, and the verbatim templates appear in the row manifests. Both surfaces keep the transition table, depth, intermediate states, and answer of the symbolic family in Section 4.1, so only the rendering changes while the computation is preserved. The sentences are templated and distractor-free, which is the sense in which the transfer is controlled. At step 6,000:

\begin{table}[!tb]
\centering
\caption{Natural-surface accuracy at step 6{,}000, full-block keeper.}
\label{tab:natural}
\small
\begin{tabular}{lrr}
\toprule
Surface & Correct / 1,536 & Accuracy \\
\midrule
Relay & \texttt{1,321} & \texttt{86.0\%} \\
Pointer & \texttt{1,213} & \texttt{79.0\%} \\
\bottomrule
\end{tabular}
\end{table}

Performance was high at shallow and middle depths but declined in the tail. Relay held 82.8\% at depth 10, then fell to 57.0\% and 32.0\% at depths 11 and 12. Pointer held 77.3\% at depth 9, then fell to 55.5\%, 44.5\%, and 19.5\% at depths 10 through 12. The checkpoint curve adds a finding. The worst-case deep-tail minimum across the two surfaces peaked at step 2,000 at 54.69\% and contracted monotonically thereafter, to 38.28\% at step 4,000 and 19.53\% at step 6,000, even as aggregate active accuracy rose from 90.12\% to 94.28\% on relay (the diagonal and aggregate figures use different matrix cells and denominators and are not directly comparable). This is the opposite of the synthetic dose pattern: surface fine-tuning past its alignment optimum improved aggregate accuracy while eroding the worst-case deep tail. The step-2,000 checkpoint, which preserved that worst-case deep tail, was therefore frozen as the natural keeper used in Section 12, and step 6,000 is reported here as the strongest checkpoint within the trained support.

These surfaces show the transition is not confined to one symbolic rendering. They are nevertheless controlled generated tasks, not broad natural reasoning benchmarks. The surface experiments also showed that retrieval organization depends on training history, since head-level specialization observed on the consolidated N24 checkpoint did not replicate on the backward-recovery checkpoint, a keeper with a different training history. A zero-shot control separates what verbal training contributes from what the installed mechanism contributes on its own. Evaluated on these frozen surfaces without any verbal training, the adapter arm scored 16.2\% on relay (249/1,536) and 17.2\% on pointer (264/1,536) against the verbally trained full-block keeper\textquotesingle s 86.0\% and 79.0\%. The full-block installation scored similarly before its own verbal training, 17.4\% on relay (267/1,536) and 20.3\% on pointer (311/1,536). Verbal competence therefore comes from verbal training, and minimal zero-shot transfer held at both budgets. What the installed mechanism does contribute once verbal training begins is measured next.

\hypertarget{matched-verbal-training-at-the-adapter-budget}{%
\subsection{Matched verbal training at the adapter budget}\label{matched-verbal-training-at-the-adapter-budget}}

The zero-shot result left open whether the installed mechanism helps at all once verbal training begins. A registered continuation answered this with two arms at the identical rank-16-plus-bridge budget, differing only in initialization history: one began from the installed adapter checkpoint, and one began from a fresh surgery on the pretrained base. Both trained on the same mix of 2,048 verbal relay rows and 2,048 synthetic rehearsal rows, with per-loop labels active and the pretrained backbone frozen throughout. Pointer was excluded from training and served as a held-out surface.

Installed history accelerated learning at every matched checkpoint. The p values are from the two-sided exact McNemar test of Section 4.2, which pairs the two arms row by row and tests whether the rows they disagree on split evenly:

\begin{table}[!tb]
\centering
\caption{Matched verbal training at the adapter budget: installed versus fresh initialization.}
\label{tab:e3b}
\small
\resizebox{\linewidth}{!}{%
\begin{tabular}{rrrrr}
\toprule
Step & Installed arm & Fresh arm & Difference & Two-sided exact p \\
\midrule
0 & 513/3,072 (16.70\%) & 86/3,072 (2.80\%) & +13.90 pp & 3.67e-82 \\
1,000 & 1,012/3,072 (32.94\%) & 571/3,072 (18.59\%) & +14.36 pp & 3.76e-56 \\
2,000 & 1,753/3,072 (57.06\%) & 697/3,072 (22.69\%) & +34.38 pp & 1.26e-210 \\
3,000 & 1,852/3,072 (60.29\%) & 1,282/3,072 (41.73\%) & +18.55 pp & 9.74e-81 \\
\bottomrule
\end{tabular}
}
\end{table}

At the last matched checkpoint the advantage was 18.6 percentage points, with 763 rows that only the installed arm answered correctly against 193 for the fresh control. The gain was not an artifact of output formatting. The two arms were essentially tied at depths 1 and 2, the advantage was concentrated at depths 3 through 11 and largest at depths 6 through 8, and it appeared at least as strongly on the held-out pointer family (917/1,536 versus 617/1,536) as on the trained relay surface (935/1,536 versus 665/1,536).

The planned 6,000-step comparison could not be completed. At step 3,000 the fresh arm\textquotesingle s Tier-1 arithmetic check moved from 60/64 to 58/64, one row beyond its three-point boundary, and the run stopped as the pre-specified regression test protocol required. Two changed rows carry no statistical weight (exact paired p of 0.50), so the stop reflects a strict rule applied to a small sample rather than demonstrated damage, but step 3,000 remains the last checkpoint at which the two arms can be compared as pre-specified rather than post hoc. Training on alone, the installed arm reached 71.45\% at step 4,000 and fell back to 64.68\% by step 6,000, below the full-block verbal references of 86.0\% and 79.0\%. Retention held throughout: with 50\% synthetic rehearsal in the training mix, every synthetic depth stratum stayed at or above 94.5\%, and the installed arm\textquotesingle s Tier-1 arithmetic check never fell below 59/64.

Thus, at a frozen adapter budget, installing the symbolic mechanism first made related verbal tasks substantially faster to learn, including a surface excluded from training, and rehearsal preserved the installed operation while the new one was acquired. The result shows that the installed mechanism transfers. It does not show parity with full-block verbal training, and it does not show that the advantage over fresh training would have survived to the planned endpoint.

\hypertarget{the-acquisition-retention-boundary}{%
\section{The Acquisition-Retention Boundary}\label{the-acquisition-retention-boundary}}

The three inverse-task branches that follow test the other side of the boundary: whether the full-block system can acquire a second, non-native operation while keeping the first, the classic catastrophic-interference setting \citep{mccloskey1989,kirkpatrick2017}. The inverse operation was chosen deliberately because it reverses the per-step retrieval. Reading a rendered table forward, the query matches the left side of a line and the answer is read from the right, the native attention pattern of a pretrained causal transformer. Reverse lookup must find the query on the \emph{right} side of a line and emit its left, a configuration related to, though distinct from, the reversal curse documented for weight-stored relations \citep{berglund2023}. Because the table is presented in context rather than stored in weights, these branches measure retrieval direction under iteration and support no claim about the reversal curse itself. The two renderings were trained at matched dose, so comparing them isolates where the difficulty lies. When the inverse relation is rendered explicitly, so each step becomes a forward-style lookup, the operation is acquired at near ceiling (Section 11.2). When the forward table must be searched in reverse, composition stalls (Section 11.1). The difficulty is the retrieval direction, not the iterative composition.

\hypertarget{canonical-forward-table-inverse}{%
\subsection{Canonical forward-table inverse}\label{canonical-forward-table-inverse}}

This branch kept the canonical forward table, so each step required the reverse lookup described above. Several curricula stalled at matched dose. Active-loop analysis revealed partial gains at loops 2 and 3 that final-answer scoring had hidden, but later loops remained unsupported. The stall was therefore not a complete failure of optimization but an incomplete staircase, and it supplied the program\textquotesingle s dose-accounting unit, active weighted labels per loop, under which the learned staircase tracked the exposure gradient exactly.

\hypertarget{explicit-inverse-table-branch}{%
\subsection{Explicit inverse-table branch}\label{explicit-inverse-table-branch}}

This branch rendered the inverse relation explicitly, so each step becomes the forward-style lookup described above. Trained in isolation, the new operation was acquired at near ceiling, 63/64. The failure appeared on the retention side. A continuation that rehearsed the original task while training the new operation brought the new operation to ceiling, 64/64, and held synthetic retention above its required floor, 0.96875 against 0.93, but natural-surface accuracy fell by 5.86 points, beyond the prespecified 3-point hard-stop margin and indicative of potential regression on core language benchmarks. The rehearsal sweep traded acquisition against retention, and no checkpoint along it satisfied every retention constraint at once.

\hypertarget{the-inverse-rendered-continuation}{%
\subsection{The inverse-rendered continuation}\label{the-inverse-rendered-continuation}}

A third branch carried the inverse-rendered operation through its calibration measurement. Calibration reached 276/384, below the preregistered pooled gate of 288/384, and the depth-4 gate failed separately at 38/96. An earlier draft and its closeout table had incorrectly reported the gate value as the achieved count. A source audit additionally placed synthetic retention below its required floor, 0.8125 against 0.93. A registered continuation worsened every relevant measure:

\begin{table}[!tb]
\centering
\caption{The inverse-rendered continuation worsened every relevant measure.}
\label{tab:inverse}
\small
\begin{tabular}{lrr}
\toprule
Measure & Before & After \\
\midrule
Calibration & \path{276/384} & \path{208/384} \\
Synthetic retention minimum & \texttt{0.8125} & \texttt{0.125} \\
Natural-surface check & \path{227/256} reference & \path{171/256} \\
\bottomrule
\end{tabular}
\end{table}

Together, these branches separate learning from retention. The operation is representable and trainable, but the tested full-block adaptation regime cannot preserve both old and new behavior on the 0.5B model substrate. Whether the adapter budget, whose frozen pretrained base confines interference to the adapter itself, meets the same boundary was then measured directly (Section 11.4).

\hypertarget{the-retention-probe-at-the-adapter-budget}{%
\subsection{The retention probe at the adapter budget}\label{the-retention-probe-at-the-adapter-budget}}

The explicit inverse-table operation was trained starting from the adapter arm\textquotesingle s final checkpoint in order to assess the acquisition-retention boundary on this smaller training substrate, with a 25-33 percent forward-rehearsal mix and every guardrail check active. One rule was specific to this arm. Because it never received verbal training, the keeper lineage\textquotesingle s references do not apply, so the natural-surface check hard-stops against the arm\textquotesingle s own pre-measured baseline of 60/256. The run hard-stopped at step 100 with the boundary intact. Inverse acquisition reached only 2/64 against the 46/64 requirement, the synthetic forward guardrail\textquotesingle s minimum stratum collapsed to 0.09375 against the 0.93 floor, and the natural-surface check fell to 49/256, a drop of 4.30 points past the 3-point tolerance. Only the Tier-1 arithmetic check passed, 59/64 against its 60/64 reference. No checkpoint achieved a joint pass.

The result replicates the boundary of Sections 11.1 through 11.3 at the adapter budget and answers the hypothesis the probe was designed to test. Confining trainable capacity to a frozen-base adapter does not confine the interference, because the adapter must host the consolidated forward mechanism and the new operation at once. The same joint-failure pattern occurred at both budgets, here with the forward mechanism collapsing within 100 steps, though the schedules were not matched, so no speed comparison is made. What the adapter budget guarantees is recovery rather than coexistence: the pretrained base is untouched and the damaged adapter is detachable. Per-operation modular adapters, one adapter per operation over a shared frozen base, are the natural successor design this result motivates.

\hypertarget{boundary-findings-and-future-directions}{%
\section{Boundary Findings and Future Directions}\label{boundary-findings-and-future-directions}}

The remaining registered measurements mark boundaries for future study rather than results this paper builds on. Each closed a line of investigation, and each points to the study that should reopen it.

\textbf{Learned depth selection.} Every mechanism claim in this paper uses forced or externally specified depth, because the existing controller could not learn the choice. Over the frozen N24 mechanism, whose forced-depth accuracy is 759/768 (98.83\%) on held-out rows, a halting head trained with supervised cross-entropy to the stated depth selected correctly on only 70/768 rows (9.11\%), and the original PonderNet objective, with no depth supervision, collapsed to always choosing the maximum depth (64/768, exactly 1/12, with zero Spearman correlation to the true depth). A third read on the adapter arm collapsed to a different single depth, so the failures differ in direction but share a kind: one global depth rather than a per-row choice. The cause is structural. Over a frozen executor that emits the correct intermediate at every loop, the outcome-weighted objective reduces to the classification problem the supervised head already failed, and the mean-pooled information path plausibly never carries total depth at all, since training consumed that value and the mechanism needed only its per-step contract. Depth control is closed as a bounded negative for this controller. A redesigned information path, and a task family whose depth is determined by content rather than stated, are the subjects of a follow-up study.

\textbf{Multi-channel re-entry.} A preregistered precursor battery asked whether the bridge should be split into learned channels. Activation required two of three positive measurements. Subspace drift was statistically distinguishable from random partitions yet far below the registered concentration threshold on both checkpoints, and retrieval-head concentration was positive on N24 but did not replicate on the backward-recovery checkpoint, so the gate was unsatisfiable and the intervention was closed without training it. The surviving finding is that retrieval structure is training-history specific.

\textbf{Loop-position transfer.} A disposable measurement asked whether the inverse operation trained at loop positions 1 and 2 transfers to positions 3 and 4. It cleared its trained-position prerequisite, 46/64 at position 1 and 64/64 at position 2, but the frozen synthetic guardrail fell to 0.8125, below the 0.93 floor, before the transfer positions were measured, and the runner deleted both disposable checkpoints as required. The result is inconclusive. It does demonstrate that the lineage policy worked: a potentially damaging branch produced no successor checkpoint.

\textbf{The width-substrate gate.} The branching-relations screen tested the prerequisite for the companion stochastic-width study: whether a frozen keeper emits valid members of an exact reachable set when several answers are correct. The gate required pooled validity of at least 0.70 with every depth at least 0.55. The natural step-2,000 keeper passed at 389/512 (75.98\%) with a minimum depth accuracy of 62.5\%, and the consolidated N24 keeper missed at 355/512 (69.34\%) with depth 3 at 52.34\%. One passing keeper is sufficient, so the guided-width study proceeds on this substrate as a separate registered companion. The screen establishes substrate competence, not useful stochastic width, and no latent prior or posterior head was trained in this program. The early unguided-width exploration is reviewed in Appendix D.

\hypertarget{discussion}{%
\section{Discussion}\label{discussion}}

\textbf{What the model learned.} The combined evidence is inconsistent with a pure final-label lookup account. The model predicts registered intermediate states, continues the transition beyond the target state, retains the operator after intermediate losses are removed, improves its depth frontier when support is increased at the frontier-to-support ratio of Section 8.1, reads out losslessly under the aligned reader, and separates from a serialized dense scratchpad most sharply beyond the scratchpad\textquotesingle s horizon.

\textbf{Why the mechanism finding matters beyond this task.} A recurring concern about pretrained language models is that adaptation installs retrievable content more readily than reusable procedure, and that stored procedural knowledge is hard to invoke reliably. That proposition has partial support in the pretraining-attribution literature, where documents influential for reasoning tend to carry reusable procedural content rather than retrieved answers, though no strict dichotomy is established \citep{ruis2024}. In a controlled setting, this study is a constructive counterpoint: an architecture, one block reused over depth, and a training regimen, per-step supervision followed by scaffold removal, that together instill a procedure the model then applies, step by step, past its trained depth and across task surfaces. The installation half of that claim now holds at two budgets, the same regimen installing the procedure through a 6M adapter over frozen weights (Section 6.2). The matched comparison further separates the requirements, installation and near-horizon competence at the adapter budget, the full-block arm stronger at depths 12-14 (Section 9.4). The boundary results sharpen rather than weaken the point: the difficulty encountered was never teaching a procedure but rather retaining several at once.

\textbf{What remains confounded.} The recurrent arm had a longer and different training lineage than the dense controls, and receives forced loop compute proportional to depth. The result establishes that this recurrently trained system realizes the task family more effectively than the evaluated dense recipes. It does not isolate weight tying, supervision format, optimization history, or compute as the sole cause.

\textbf{Why the retention boundary matters.} The inverse experiments show a converted pretrained model under full-block adaptation is not an empty substrate: new training competes with a consolidated transition and with retained language-model behavior. Larger substrates, detachable adapters, routing, or explicit modularity may move this boundary. Weakening guardrails would not answer the scientific question, because it would trade one capability for another.

\hypertarget{limitations}{%
\section{Limitations}\label{limitations}}

These limitations bound every claim above.

\begin{itemize}
\tightlist
\item
  Results center on Qwen2.5-0.5B and controlled synthetic or generated surfaces.
\item
  Phase A does not match training tokens, optimizer history, FLOPs, latency, or inference compute.
\item
  The N24 depth frontier is a single-seed result. Seed variance in consolidation speed was measured at a smaller support. The adapter-budget arm is likewise a single training seed.
\item
  The adapter arm\textquotesingle s far-tail deficit is not attributed among rank, frozen-base geometry, bridge capacity, and optimization. No rank sweep was run.
\item
  Zero-shot verbal transfer is minimal at both budgets\textquotesingle{} baselines. Adapter-budget verbal training is a single seed on controlled generated tasks, with the pointer surface held out from training but drawn from a related generator family, synthetic retention measured under a 50\% rehearsal mix, and the fresh-control arm truncated by its guardrail before the planned endpoint.
\item
  Installation order was not varied: the retention boundary is characterized for forward-first installation only, and inverse-first installation is untested.
\item
  The adapter retention probe is a single seed and a single rehearsal configuration. Per-operation modular adapters were not tested.
\item
  Forced-depth evaluation measures the transition mechanism separately from learned depth selection. The bounded selector experiment closed the existing controller as a negative (Section 12).
\item
  Full-block unfreezing uses a substantial 182M-parameter adaptation budget.
\item
  External benchmarks such as GPQA Diamond, ARC-AGI, mathematics, and coding have not established a superiority claim. The preservation result of Section 7 is non-inferiority on the evaluated battery, not universal capability preservation.
\item
  The natural surfaces are controlled renderings and do not substitute for broad language understanding.
\item
  The retention boundary replicated at both tested budgets, but it may still move with scale, routing, per-operation modular adapters, or a different curriculum.
\item
  Guided stochastic width is evaluated in a separate registered study. This paper makes no width claim in either direction.
\end{itemize}

\hypertarget{reproducibility-and-artifact-map}{%
\section{Reproducibility and Artifact Map}\label{reproducibility-and-artifact-map}}

The public repository carries the full evidence chain under outputs/ and docs/ at \url{https://github.com/mshapiro123/recurrent-qwen-svgd}. The machine-readable claim ledger, \path{part1_claim_evidence_ledger.json}, links every claim in this paper to its evidence record (Section 4.3). Per-result summary.json files record row-level predictions and RNG manifests for every reported experiment, covering the persistent chain, the N24 frontier, the frozen depth-22 matrix, Phase A, natural transfer, the inverse branches, the bridge battery, the adapter arm with its E2, E3a, E3b, and E4 batteries, the guardrail battery, the frontier threshold-sensitivity check, and the wall-clock latency run. The figures are under docs/figures/, and the preregistration documents ship with their locked thresholds. The full-block natural-surface retention canary is preregistered in \path{docs/INVERSE_COMPOSITION_STAIRCASE_SPEC.md} under Safety and Readings, and the adapter-lineage natural-surface check is preregistered in \path{docs/ARM_E_ADAPTER_PARITY_BATTERY_SPEC_20260719.md} under E4. Experiments used Google Colab GPU runtimes, with accelerator type and per-run timing reported where preserved in individual receipts. Session-level runtime records were not retained consistently, including exploratory and failed debugging runs, so a defensible total GPU-hour figure cannot be reconstructed. No aggregate compute figure is therefore reported.

\hypertarget{conclusion}{%
\section{Conclusion}\label{conclusion}}

A pretrained Qwen2.5-0.5B model can be converted into an identity-preserving recurrent-depth architecture, and the conversion succeeds at two measured parameter budgets, a 182M full-block adaptation and a 6M adapter over frozen base weights. The converted model learns a persistent loop-indexed state transition whose validity frontier scales at 1.44 to 1.50 times its supervised support. On the frozen synthetic family, evaluated under a lossless aligned reader, it outperforms the registered dense recipes overall, and most decisively beyond the serialized scratchpad\textquotesingle s learned horizon.

The same program measured the limits of that result. General natural reasoning remains unproven. Learned depth selection is a registered bounded negative for the current controller. And neither budget could retain a newly acquired inverse operation alongside the consolidated mechanism (Section 11.4).

The deterministic program is complete. The keeper checkpoints are frozen, the branching-relations substrate has passed its deterministic validity gate, and one registered question remains for the companion study: whether target-conditioned stochastic latent trajectories add exact multi-solution coverage beyond output sampling and beyond additional depth.

\section*{Acknowledgments}
This work was self-funded by the author. No external funding was received.

\appendix
\hypertarget{negative-results-from-the-initial-parameter-efficient-attempt}{%
\section{Negative Results from the Initial Parameter-Efficient Attempt}\label{negative-results-from-the-initial-parameter-efficient-attempt}}

\hypertarget{the-original-question-and-design}{%
\subsection{The original question and design}\label{the-original-question-and-design}}

The project\textquotesingle s first question was narrower than this paper\textquotesingle s: can a pretrained model be retrofitted into a recurrent-depth system without updating its pretrained base weights, using only small trainable additions? The initial design held every pretrained weight frozen and trained three components: LoRA adapters inside the looped block; a re-entry bridge initialized to identity with a zero-initialized mixing gate, so that training would begin from exact base behavior; and a PonderNet-style sequence-level halting head over loop counts, regularized toward a centered geometric prior, intended to learn how much recurrent depth a problem requires. The recorded configuration: rank-8 LoRA (alpha 16) on all seven projection families across the twelve block layers, 84 modules; a Ponder head with initial halt probability 0.15 and KL weight 0.08; and, in its second phase, four SVGD particles over a 256-dimensional latent. The era predated the registered gate framework. Its 128-row screening recovered 70/128 (Phase 1) and 69/128 (Phase 2) against a 72/128 base, historical outcomes rather than gate-qualified evidence.

\hypertarget{why-it-failed-decomposed}{%
\subsection{Why it failed, decomposed}\label{why-it-failed-decomposed}}

The failure was not one thing, and most of its components would have defeated any parameter budget.

The first defect was the loop closure. The era\textquotesingle s iteration omitted the Prelude re-injection entirely. Without re-injected context, iterated states leave the trained manifold (Section 1), so every adapter-era recurrence result was computed on a loop that was not the intended system. That confound cannot be removed retroactively: the era\textquotesingle s negatives are evidence about a malformed loop with small trainable additions, not about parameter-efficiency itself. The production architecture\textquotesingle s closure audits are those of Section 3.2.

The second defect was a self-locked gate. The identity-preserving initialization contained its own trap: with the mixing gate at zero, the bridge contributed nothing to the forward pass, so it received no useful gradient, so the gate stayed at zero. Diagnostics on the relevant checkpoints found a zero gate, zero bridge output delta, and gate, weight, and bias gradients all exactly zero. A read-only control forcing the gate to one immediately recovered nonzero projection gradients (weight RMS 9.28e-4), isolating dead initialization rather than any optimizer failure. The pathway that was supposed to learn the recurrent combination was inactive from the first step. The repaired program replaced this with the identity-biased but live gate of Section 3.2 and verified liveness explicitly (gate movement to \textasciitilde1.0002, live gradients, bounded loop-output RMS). The supported lesson is narrower than a general rule: the paired zero-gate, zero-branch initialization used here was self-locking. Liveness must be asserted, not assumed.

Under these conditions the halting controller had nothing meaningful to allocate depth over. Recorded telemetry shows expected loop counts of 2.69 to 3.07 with halting entropy near 1.3 across the era\textquotesingle s readouts, a broad, task-independent stopping distribution, neither collapsed nor adaptive, directly consistent with the bounded selector negative later measured over a working mechanism (Section 12).

After the structural repairs, the program moved to full-block adaptation, which succeeded and produced the reference installation and the deep characterization. The controlled adapter arms on the repaired loop were then run as registered experiments, the rank-16 closure (Section 6.2) and the matched adapter arm (Section 9.4). Together they answer the question the era could not. The mechanism installs parameter-efficiently on a sound architecture, so the original failure is attributable to the broken machine, not the budget.

\hypertarget{the-bounded-answer}{%
\subsection{The bounded answer}\label{the-bounded-answer}}

For every configuration tested in the original era, the answer to the opening question was no, and two structural defects would have defeated any budget. The registered closure experiment of Section 6.2 asked the question again on the corrected architecture, and the answer inverted: the mechanism installs at a 6.01M forward-active adapter budget with the base weights untouched, bounded exactly as Section 6.3 states it. What the frozen base guarantees is \emph{recovery}. Section 11.4 shows the single small adapter did not host a second operation, so hosting multiple operations plausibly requires one adapter per operation over the shared frozen base. One methodological note from the matched comparison belongs here as well: pooling accuracy across depths hides the Section 9.4 crossover entirely, which is why the registered gate was the per-depth profile and why depth-stratified reporting is used throughout.

\hypertarget{the-retired-stability-damper}{%
\subsection{The retired stability damper}\label{the-retired-stability-damper}}

The stability era introduced a per-axis damper on the loop state, a per-dimension attenuation intended to bound iterate growth while the loop closure was still defective. After the architectural repair, the mechanism proved unnecessary: the production path carries it at strength 0.0, and Section 3.5 notes its dormancy only so repository readers do not mistake inactive code for an active mechanism. The damper was an eval-only, covariance-calibrated principal-component tail attenuation: at full strength it reduced the loop-8 tail trace ratio from 33.65 to 2.91 on a 512-row ARC-Challenge sweep without a clean answer-selection gain (52 rescued versus 67 harmed), and it was retired. A recorded correspondence with Parcae \citep{prairie2026} did not survive an equivalence check and stands only as a high-level stability analogy. The thematic point stands: the broken loop needed external stabilization, and the corrected one does not, which is itself a small verification that the repair addressed the cause rather than the symptom.

\hypertarget{latency-decomposition}{%
\section{Latency Decomposition}\label{latency-decomposition}}

The primary latency table in Section 9.5 reports complete synchronized model calls. For decomposition, each family was also measured against a synchronized minimum-path reference. Recurrent incremental time is the forced-depth call minus a one-loop reference, and dense incremental time is the full registered generation call minus a one-token generation reference. Both families therefore use baseline subtraction, with different reference operations. The overall reference medians were 28.9 ms for the full-block recurrent call, 42.3 ms for the adapter recurrent call, and 27.7, 27.9, and 31.8 ms for the scratchpad, direct 0.5B, and direct 1.5B one-token references, varying by less than approximately 0.2 ms across the displayed depths.

The incremental medians per row (ms), the additional model-side time beyond the synchronized one-step reference. Each cell is the median of per-row differences, so it need not equal the difference of the two medians:

\begin{table}[!tb]
\centering
\caption{Baseline-subtracted incremental latency medians per row (ms).}
\label{tab:decomp}
\small
\resizebox{\linewidth}{!}{%
\begin{tabular}{rrrrrr}
\toprule
Depth & Full-block recurrent & Adapter recurrent & Scratchpad & Direct 0.5B & Direct 1.5B \\
\midrule
1 & 0.0 & 0.0 & 2,259 & 742 & 873 \\
2 & 20.9 & 34.2 & 2,262 & 742 & 875 \\
4 & 62.7 & 102.8 & 2,260 & 744 & 873 \\
8 & 147.0 & 240.1 & 2,256 & 741 & 872 \\
11 & 209.7 & 342.9 & 2,249 & 740 & 874 \\
14 & 272.9 & 445.9 & 2,256 & 742 & 872 \\
\bottomrule
\end{tabular}
}
\end{table}

\hypertarget{the-dense-arm-reader}{%
\section{The Dense-Arm Reader}\label{the-dense-arm-reader}}

The reader used for the dense arms in Section 9 was corrected after a post-closeout audit. The original reader selected the last answer marker in a generation. The dense models were not trained to emit an end-of-sequence token, often produced a correct answer and then continued generating, and the continuation overwrote the completed answer. The corrected reader accepts the completed answer regardless of any text that follows it, which is generous to the dense arms. The recurrent arms read a single decoded state and were unaffected by the audit. All dense-arm numbers in Section 9 use the corrected reader, and the audit receipts are in the public repository.

\hypertarget{early-stochastic-width-exploration}{%
\section{Early Stochastic-Width Exploration}\label{early-stochastic-width-exploration}}

Early development runs explored stochastic width before the recurrence was repaired. Injected noise and SVGD-style repulsion increased candidate diversity in some settings but did not reliably produce new correct candidates. Those runs also lacked the target-conditioned variational guidance that distinguishes GRAM-style approaches from unguided noise, so they answer the width question in neither direction. Guided stochastic width remains open. The registered companion study will freeze the deterministic keeper, train only prior and posterior latent heads with an injection scale, and compare exact coverage at matched K against answer-head sampling and matched transition compute.

\hypertarget{optimizer-update-cancellation}{%
\section{Optimizer Update Cancellation}\label{optimizer-update-cancellation}}

Section 5.1 reports that gradient-slice scaling inside a combined projection fails to deliver a raised learning rate under either optimizer family. The derivation follows. Under Adam-family updates, a single parameter moves by

\[\Delta\theta_t = -\eta\,\frac{\hat m_t}{\sqrt{\hat v_t}+\epsilon}, \qquad m_t=\beta_1 m_{t-1}+(1-\beta_1)\,g_t, \qquad v_t=\beta_2 v_{t-1}+(1-\beta_2)\,g_t^2.\]

Scaling the gradient by a fixed factor \(c>0\) on every step scales the moment estimates to \(c\,\hat m_t\) and \(c^2\,\hat v_t\), and the factor cancels in the ratio,

\[-\eta\,\frac{c\,\hat m_t}{\sqrt{c^2\,\hat v_t}+\epsilon} \;=\; -\eta\,\frac{\hat m_t}{\sqrt{\hat v_t}+\epsilon/c},\]

surviving only inside the epsilon term, so the realized update is unchanged up to the epsilon floor. The cancellation is exact once the moment buffers reflect the scaled gradient history, so changing the factor mid-training produces only a transient. AdamW\textquotesingle s decoupled weight decay enters outside the ratio and is untouched by gradient scaling in any case.

Under Muon, the update applies SGD-style momentum and then orthogonalizes the momentum matrix toward its polar factor by Newton-Schulz iteration \citep{jordan2024}, so with \(M_t = U\Sigma V^\top\) the applied update is approximately \(-\eta\,UV^\top\). The polar map satisfies \(\operatorname{polar}(cM_t) = \operatorname{polar}(M_t)\) for \(c>0\), so a global gradient scale is discarded by the exact polar map, and in practice up to the iteration\textquotesingle s norm epsilon and numerical error. A factor applied to one slice of the combined matrix is not a global scale. It generally changes the singular vectors, and even where only the singular values change, the orthogonalized update does not preserve the slice-local multiplier.

\end{document}